\documentclass[10pt, journal]{IEEEtran}
\usepackage{array}
\usepackage{graphicx}
\usepackage{amsmath}
\usepackage{amssymb}
\usepackage{bm}
\usepackage{booktabs}
\usepackage{pifont}
\usepackage{adjustbox}
\usepackage{multirow}
\usepackage{makecell}
\usepackage{url}
\usepackage{xcolor}
\usepackage{subfigure}
\usepackage[colorlinks=false, pdfborder={0 0 0}]{hyperref}

\usepackage{newtxtext}

\let\oldurl\url
\renewcommand{\url}[1]{\textcolor{green!50!black}{\oldurl{#1}}}

\usepackage{xcolor}

\newcolumntype{C}[1]{>{\centering\arraybackslash}p{#1}}

\makeatletter
\newcommand{\thickhline}{%
	\noalign {\ifnum 0=`}\fi \hrule height 1pt
	\futurelet \reserved@a \@xhline
}
\newcolumntype{"}{@{\vrule width 1pt}}
\makeatother

\ifCLASSINFOpdf
\else
\fi

\begin{document}
	
	\title{Fast Inference of Visual Autoregressive Model with Adjacency-Adaptive Dynamical Draft Trees}
	
	\author{Haodong Lei, Hongsong Wang, Xin Geng, Liang Wang,~\IEEEmembership{Fellow, IEEE}, and Pan Zhou
		\IEEEcompsocitemizethanks{
			\IEEEcompsocthanksitem This work was supported by National Science Foundation of China (62172090, 52441503), Jiangsu Province Natural Science Fund (BK20230833), the Fundamental Research Funds for the Central Universities (2242025K30024) and the Open Research Fund of the State Key Laboratory of Multimodal Artificial Intelligence Systems (E5SP060116). We thank the Big Data Computing Center of Southeast University for providing the facility support on the numerical calculations. 
			\IEEEcompsocthanksitem H. Lei, H. Wang and X. Geng are with School of Computer Science and Engineering, Southeast University, Nanjing 211189, China, and also with Key Laboratory of New Generation Artificial Intelligence Technology and Its Interdisciplinary Applications (Southeast University), Ministry of Education, China (email: leihaodong@seu.edu.cn;hongsongwang@seu.edu.cn;xgeng@seu.edu.cn).
			\IEEEcompsocthanksitem L. Wang is with New Laboratory of Pattern Recognition (NLPR), State Key Laboratory of Multimodal Artificial Intelligence Systems (MAIS), Institute of Automation, Chinese Academy of Sciences (CASIA), and also with School of Artificial Intelligence, University of Chinese Academy of Sciences (email: wangliang@nlpr.ia.ac.cn).
			\IEEEcompsocthanksitem P. Zhou is with School of Computing and Information Systems, Singapore Management University (email: panzhou@smu.edu.sg).
		}
	}
	
	\maketitle
	
	\begin{abstract}
		Autoregressive (AR) image models achieve diffusion-level quality but suffer from sequential inference, requiring approximately 2,000 steps for a 576×576 image. Speculative decoding with draft trees accelerates LLMs yet underperforms on visual AR models due to spatially varying token prediction difficulty. We identify a key obstacle in applying speculative decoding to visual AR models: inconsistent acceptance rates across draft trees due to varying prediction difficulties in different image regions. We propose Adjacency-Adaptive Dynamical Draft Trees (ADT-Tree), an adjacency-adaptive dynamic draft tree that dynamically adjusts draft tree depth and width by leveraging adjacent token states and prior acceptance rates. ADT-Tree initializes via horizontal adjacency, then refines depth/width via bisectional adaptation, yielding deeper trees in simple regions and wider trees in complex ones. The empirical evaluations on MS-COCO 2017 and PartiPrompts demonstrate that ADT-Tree achieves speedups of 3.13× and 3.05×, respectively. Moreover, it integrates seamlessly with relaxed sampling methods such as LANTERN, enabling further acceleration. Code is available at \url{https://github.com/Haodong-Lei-Ray/ADT-Tree}.
	\end{abstract}
	
	\IEEEpeerreviewmaketitle
	\section{Introduction}\label{sec:intro}
	
	The era of foundation models has ushered in a paradigm shift in visual synthesis. Autoregressive (AR) image generators, exemplified by Emu3~\cite{EMU3}, Anole~\cite{Anole}, and Lumina-mGPT~\cite{luminamgpt,luminamgpt2}, now outperform the diffusion model in terms of image generation quality~\cite{ARvsDiff,LDC,Llamagen} while inheriting the scalable pretraining recipes of large language models (LLMs). Trained on billions of image-text pairs, these visual AR models tokenize raw pixels into discrete sequences via quantized autoencoders~\cite{QGA} and predict tokens sequentially with transformer backbones. Despite achieving favorable image quality, their sequential inference remains a critical bottleneck, which is that generating a $576\!\times\!576$ image requires about \textbf{2,000} autoregressive steps~\cite{EMU3}. The speed of the AR visual model is orders of magnitude slower than diffusion counterparts that converge in 20--50 denoising steps. This constitutes one of the critical barriers to deploying visual AR models for production-grade image generation applications.
	
	Speculative decoding~\cite{SSM,SSM1} has emerged as the standard acceleration paradigm for LLMs, consistently achieving $2-4\times$ speedups via a ``draft-then-verify'' protocol in the textual generation. This naturally positions it as a highly promising approach for accelerating visual AR models. In particular, tree-structured variants, such as SpecInfer~\cite{SpecInfer}, Medusa~\cite{MEDUSA}, and EAGLE-2~\cite{EAGLE-2} further amplify gains by expanding the drafting search space.
	
	\begin{figure}[t]
		\centering
		\includegraphics[width=1\linewidth]{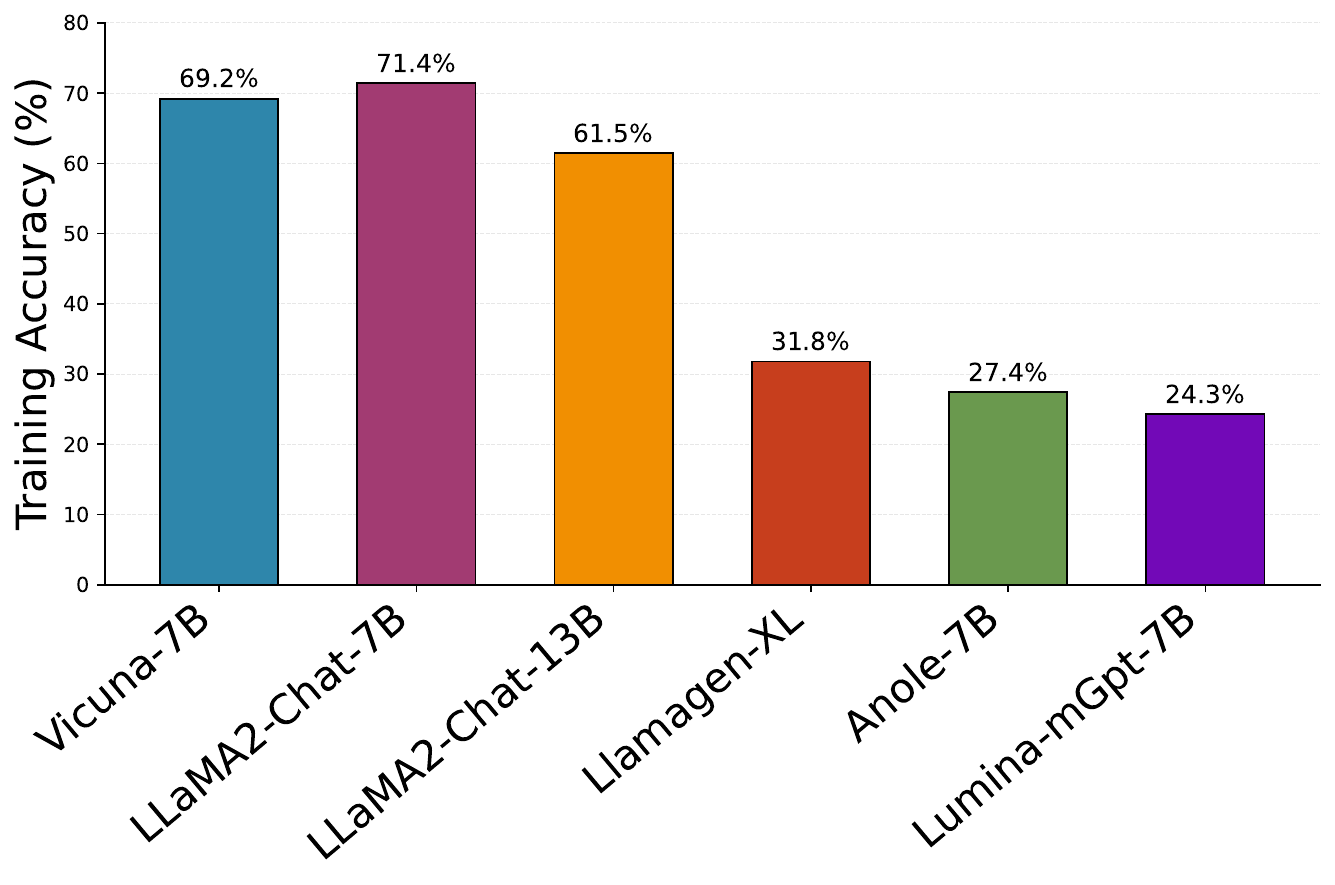}
		\caption{\textbf{The comparison of the training-set accuracy of draft models for the text-to-text and the text-to-image.} Vicuna-7B~\cite{Vicuna}, LLaMA2-Chat-7B and LLaMA2-Chat-13B~\cite{llamachat2} are text-to-text generation models. Llamagen-XL, Anole-7B~\cite{Anole}, and Lumina-mGpt-7B are text-to-image generation models. Both draft models are built upon the EAGLE draft model architecture~\cite{EAGLE-1}.}
		\label{fig:intro1}
	\end{figure}
	
	Yet, naïve transplantation of these methods to visual AR models yields unsatisfactory results. As shown in Figure~\ref{fig:intro1}, the draft model of the text-to-text Vicuna-7B~\cite{Vicuna} achieves nearly 70\% accuracy on its training set. In stark contrast, the draft model of the text-to-image Anole-7B~\cite{Anole} attains less than half of that accuracy on its corresponding training data. Existing speculative decoding methods for image generation can be broadly classified into two categories to address this challenge.
	
	The first category comprises relaxed sampling approaches, which transfer LLM speculative decoding to the visual domain by introducing relaxed sampling strategies to mitigate the severely low token acceptance rates (often below 48\%~\cite{LANTERN}) induced by visual ambiguity. For instance, LANTERN~\cite{LANTERN} modifies EAGLE-2~\cite{EAGLE-2} with relaxed verification criteria to accept a longer prefix of draft tokens, whereas LANTERN++~\cite{LANTERN++} replaces dynamic draft trees with static tree structures to ensure more stable acceleration. 
	The second category consists of self-speculative decoding paradigms, exemplified by SJD~\cite{SJD} and GSD~\cite{GSD}, which eliminate the draft model entirely to circumvent underfitting—achieving acceleration without any auxiliary training. SJD pioneers this training-free framework by leveraging the target model itself for both drafting and verification in a Jacobi-style iterative manner. GSD~\cite{GSD} further enhances stability by exploiting token clustering based on latent similarity, enabling parallel speculation over semantically redundant visual tokens.
	
	\begin{figure}[t]
		\centering
		\includegraphics[width=\linewidth]{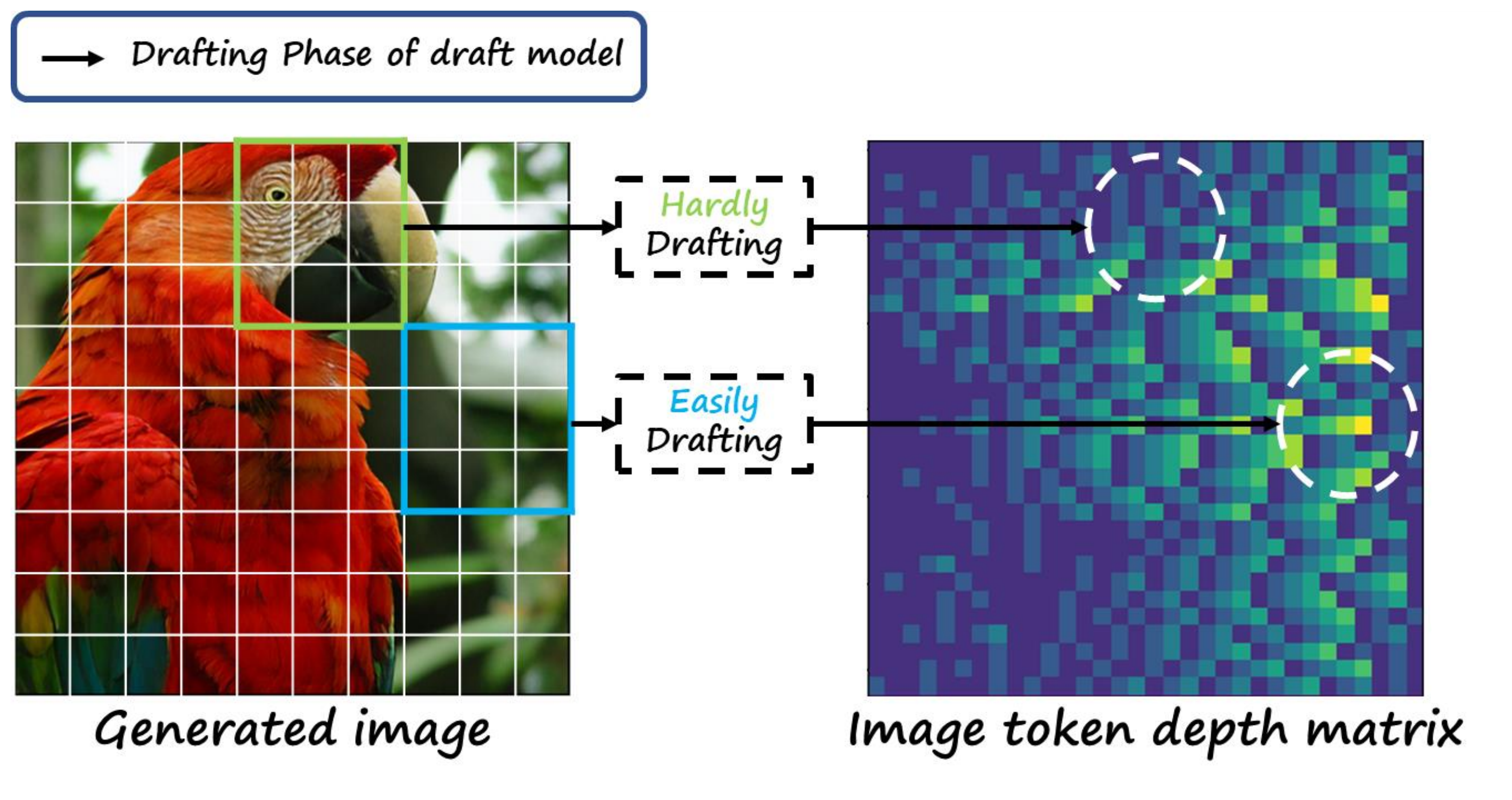}
		\caption{\textbf{The draft model faces two situations in different image regions.} The image token depth matrix tracks the depth of the draft tree at which each image token resides. In this matrix, brighter areas signify deeper locations of the image tokens within the draft tree. For complex regions, the acceptance length is lower than the height of the draft tree, making unused layers wasteful and reducing the acceleration rate. A shallow draft tree is appropriate. For simple regions, the potential acceptance length exceeds the draft tree height, so building a deeper tree can increase the acceptance length and boost the acceleration rate.}
		\label{fig:Figure1}
	\end{figure}
	
	However, while significant efforts have been devoted to comparing the differences between text and image generation under the AR model, we discover that the difference in image token generation exhibits striking spatial disparities across different regions, even within a single image. As shown in Figure~\ref{fig:Figure1}, acceptance lengths $\tau$ exhibit extreme spatial variance: smooth backgrounds yield $\tau > 3$, while textured regions collapse to $\tau \leq 2$. This imbalance renders static draft trees—whether deep-and-narrow or shallow-and-wide—fundamentally inefficient: \textbf{over-provisioning in simple texture regions wastes compute, while under-provisioning in complex texture regions caps acceleration}.
	
	To address the above issues, we propose a solution of building draft trees dubbed as ADT-Tree (\textbf{A}djacency-adaptive \textbf{D}ynamical draf\textbf{T} \textbf{Trees}) that makes use of the varying difficulty in sampling from different positions of the image to dynamically adjust the depth and top-k of the draft tree, thereby enhancing the acceptance rate and acceptance length. Specifically, we utilize the similarity between depth and probability positions of adjacent draft tokens in the draft tree to more accurately initialize the current draft tree. Then, based on the state of the previous draft trees, we adjust the expected depth and width (top-k) of draft trees through appropriate corrections. Thus, we select the depth and top-k of the draft tree more precisely to achieve a higher utilization rate of the draft tree.
	
	Our approach achieves speed-up rate raising in the speculative decoding of the token sequence, which is equipped with the characteristics of image tokens. We achieve a 3.13× speedup on MSCOCO2017~\cite{MSCOCO} and a 3.05× speedup on PartiPrompts~\cite{parti-prompts}.
	
	To summarize, our key contributions are as follows:
	\begin{itemize}
		\item 
		\textbf{Observation of the bottleneck in efficient visual speculative decoding}: We conduct extensive experiments and find that the imbalance in acceptance ratios across different image regions in the current draft tree constitutes the primary bottleneck in applying draft tree speculative decoding to visual AR models.
		\item 
		\textbf{Novel method for dynamically building draft tree}: We introduce a zero-overhead, plug-and-play algorithm that dynamically reshapes draft trees using only adjacent acceptance statistics. We design a dynamically building draft tree method, adapting the adjacent states of tokens dubbed as ADT-Tree. ADT-Tree first initializes the draft tree based on horizontally adjacent draft trees, and subsequently adjusts it according to the states of the adjacent draft trees, leading to a higher draft tree utilization rate without sacrificing image generation performance.
	\end{itemize}
	
	\section{Related Work}\label{sec:related_work}
	\noindent \textbf{Visual Autoregressive Models:} AR models~\cite{VAR,vqvae} have gained prominence in image generation~\cite{LDC}, delivering quality rivaling diffusion models~\cite{LDM-4, DDIM} through sequential token prediction. Unlike diffusion models, visual AR models tokenize images into discrete sequences and process them with transformer architectures, the same to large language models (LLMs). Existing works like LlamaGen~\cite{Llamagen}, Anole~\cite{Anole}, and Lumina-mGPT~\cite{luminamgpt} excel in text-conditional image generation, using quantized autoencoders to convert images into token sequences for transformer-based sampling.
	
	\noindent \textbf{Speculative Decoding:}
	The core idea of speculative decoding \cite{SSM,SSM1,Cascade} is to first draft and then verify: quickly generate a potentially correct draft and then check which tokens in the draft can be accepted. This method first applies to large language models with AR structure. The initial draft form is the chain structure~\cite{JD,lookhead,CLLMs}. And then SpecInfer \cite{SpecInfer} introduces a draft form with tree structure, which represents \textbf{draft tree}. The draft tree is equipped with two parameters, top-k $\hat{k}$ and depth $\hat{d}$, where $\hat{k}$ represents the number of each child node in the draft tree and $\hat{d}$ represents the depth of the draft tree. The draft form with tree structure~\cite{SpecInfer,MEDUSA,EAGLE-2,HASS} has flourished. From MEDUSA~\cite{MEDUSA} to EAGLE-2~\cite{EAGLE-2}, unleashing the potential of the tree structure draft tree, these methods greatly increase the speed-up ratio.
	
	One of the few works related to speculative decoding of image token sequences is speculative decoding for Multi-LLM \cite{SPD4MLLM}, which provides a simple yet efficient approach to applying speculative decoding in Multi-LLMs. With the introduction of Speculative Jacobi Decoding \cite{SJD}, speculative decoding has been extended to visual AR models. Although the GSD~\cite{GSD} method, based on SJD, has modified its sampling paradigm, the structure of its draft token remains a chain structure. However, the draft tokens in these methods follow a chain structure rather than a tree structure. Existing draft tree methods like LANTERN~\cite{LANTERN} employ a lossy tree-structured drafting approach with relaxation of speculative decoding.
	
	\section{Preliminaries and Motivation}
	\label{sec:Preliminaries}
	We first introduce the necessary notation. Then, we describe the motivations for ADT-Tree, highlighting the challenges and solutions for optimizing inference efficiency while maintaining the quality of conditional generation. 
	
	\subsection{Notation}
	\label{sec:Notation}
	Drawing from LLMs, we adapt speculative decoding for image generation. An image is tokenized into a sequence $ S = (s_1, s_2, \ldots, s_T) $ via a quantized autoencoder~\cite{ViT}, where a lightweight encoder and quantizer produce discrete tokens $ s_t \in \{1, \ldots, K\} $ (codebook size $ K $), and a decoder reconstructs $ \hat{I} $ from $ S $. The target model $ \mathcal{L} $, an autoregressive transformer, generates $S$ conditioned on a prompt $ \rho $ (e.g., text or label). We define $ p(s_t | s_{1:t-1}, \rho) = \mathcal{L}(s_t | s_{1:t-1}, \rho) $ as the sampling result of the conditional generation function (CFG)~\cite{CFG} for the target model. A smaller draft model $ \mathcal{R} $ generates $ q(s_t | s_{1:t-1}, \rho) $ approximates the output of $ \mathcal{L} $. In speculative decoding in visual AR models, given a prefix $ s_{1:t-1} $ and $ \rho $, $ \mathcal{R} $ proposes a draft sequence $ \hat{s}_{t+1:t+L} $ of length $ L $, which $ \mathcal{L} $ verifies in parallel. Among them, $L$ represents the total number of tokens in the draft tree. We define $\hat{s}_{ans(t)}$ as the ancestor sequence to node $\hat{s}_{t}$ based on the tree mask, which means $\hat{s}_{ans(t)}$ is the sequence from root to $ \hat{s}_{t} $. The acceptance probability is:
	\begin{align}
		r_{t+j} = \min \left( 1, \frac{p \left( \hat{s}_{t+j} | s_{1:t}, \hat{s}_{ans(t+j)}, \rho \right)}{q \left( \hat{s}_{t+j} | s_{1:t}, \hat{s}_{ans(t+j)}, \rho \right)} \right),j = 1, \ldots, L
		\label{eq:p/q}
	\end{align}
	where both $p(\hat{s}_{t+j} \mid s_{1:t-1}, \hat{s}_{\text{ans}(t+j)}, \rho)$ and $q(\hat{s}_{t+j} \mid s_{1:t-1}, \hat{s}_{\text{ans}(t+j)}, \rho)$ are computed using CFG.
	
	To further optimize drafting, we integrate a dynamic draft tree $ \mathcal{T}_{\text{draft}} $, based on EAGLE-2, having depth $ \hat{d} $ and width $ \hat{k} $. Each node $ v $ of the draft tree represents a token $ s_v $ with confidence $ c_v = q(s_v | s_{1:t-1}, s_{\text{anc}(v)}, \rho) $. The tree expands by selecting the top-$ k_d $ nodes at depth $ d $ based on path confidence $ P_v = \prod_{u \in \text{Path}(\text{root}, v)} c_u $, where $ \text{Path}(\text{root}, v) $ is the sequence from root to $ v $. For each selected node at position $ (d, k) $, $ \mathcal{R} $ generates $ k_{d+1} $ child nodes at depth $ d+1 $, positioned at $ (d+1, 1), \ldots, (d+1, k_{d+1}) $, sampling from $ q(\cdot | s_{1:t-1}, s_{\text{anc}(v)}, \rho) $, with $ k_{d+1} < \hat{k} $. The sequence is then reranked and verified with $ \mathcal{L} $.
	
	\subsection{Motivation}
	\label{sec:Motivation}
	
	\begin{figure}[t]
		\centering
		\includegraphics[width=1\linewidth]{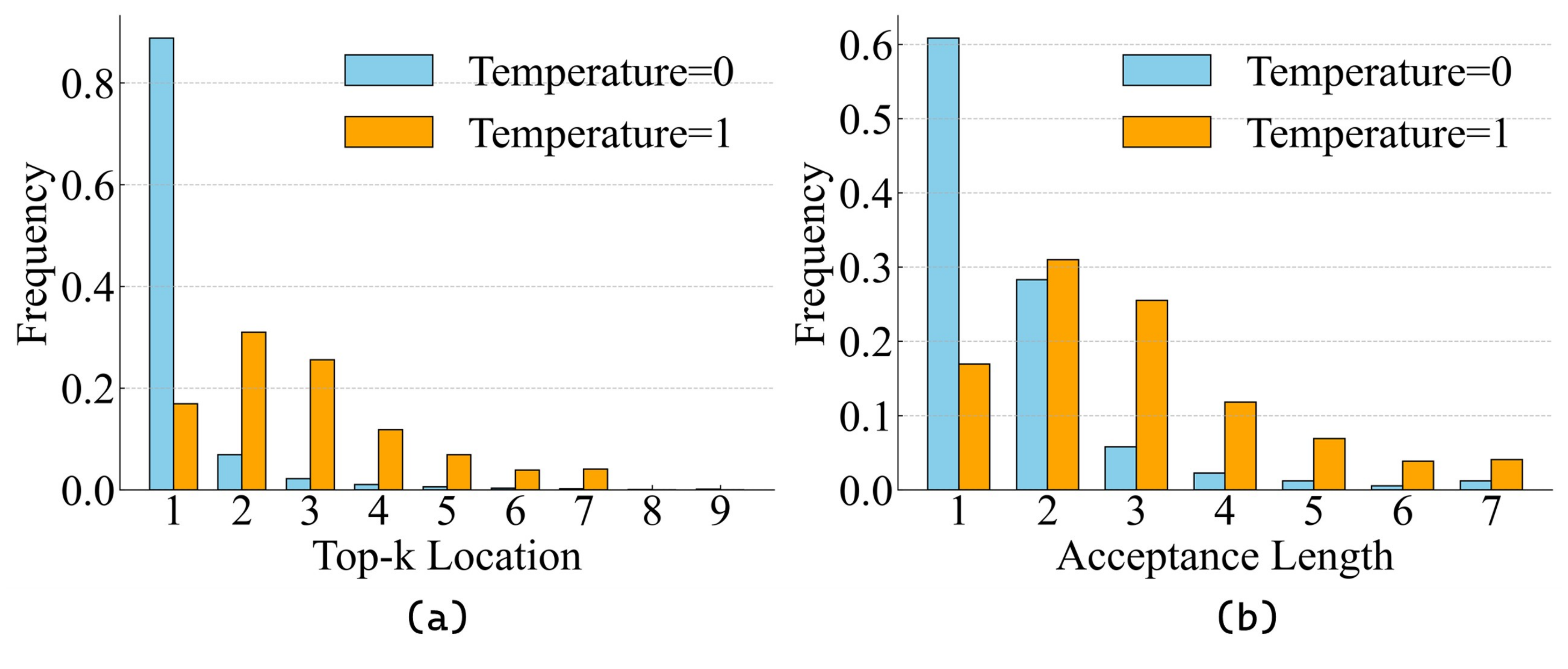}
		\caption{
			\textbf{(a) Left:} Frequency of acceptance lengths during speculative decoding with $ \mathcal{T}_{\text{draft}} $ ($ \hat{d} = 5 $, $ \hat{k} = 10 $) over 100 image generations using Anole at $ T = 0 $ and $ T = 1 $. 
			\textbf{(b) Right:} Frequency of the top-$k$ positions of accepted draft tokens, where `Top-$k$ Location' denotes the minimum $k_d$ required for $ \mathcal{R} $ to include the correct token in the draft phase for acceptance by $ \mathcal{L} $.
		}
		\label{fig:acceptance_topk_analysis}  
	\end{figure}
	
	Speculative decoding has demonstrated significant success in accelerating AR models for text generation~\cite{SSM,SSM1}. Recent advancements, such as those employing draft tree structure~\cite{SpecInfer,MEDUSA,EAGLE-2}, have expanded the search space for draft tokens. Notably, EAGLE-2~\cite{EAGLE-2} introduces a dynamic candidates token tree, $ \mathcal{T}_{\text{draft}} $, with configurable depth $ \hat{d} $ and width $ \hat{k} $, enabling manual adjustment of the token search scale. This flexibility positions EAGLE-2 as a promising approach for accelerating visual AR models ($ \mathcal{L} $), which generate token sequences $S$ conditioned on a prompt $ \rho $.
	
	However, applying EAGLE-2 to visual AR models reveals inefficiencies stemming from the expansive search scale of $ \mathcal{T}_{\text{draft}} $. To investigate this, we analyze the frequency of acceptance lengths during speculative decoding with $ \mathcal{L} $ and a draft model $ \mathcal{R} $. Figure~\ref{fig:acceptance_topk_analysis}(a) illustrates the distribution of acceptance lengths over 100 image generation trials, using a draft tree configured with $ \hat{d} = 7 $ and $ \hat{k} = 10 $, under temperature settings $ T = 0 $ and $ T = 1 $. At $ T = 1 $, the acceptance lengths exhibit significant variance, indicating that a static $ \hat{d} $ leads to inefficiencies. For instance, when the acceptance length $ \tau $ is 3, constructing a tree of depth 7 wastes computational resources on four unnecessary layers. Conversely, reducing $ \hat{d} $ to 3 caps $ \tau $ at 3, limiting the potential acceleration in regions where $ \mathcal{R} $ could predict longer sequences. This trade-off complicates the selection of an optimal $ \hat{d} $ for visual AR speculative decoding, a phenomenon also noted in prior works such as SJD~\cite{SJD} and LANTERN~\cite{LANTERN}, which highlight local similarities in token generation.
	
	We identify a critical challenge: \textbf{imbalance in acceptance rates of draft trees}. During speculative decoding of the token sequence $ S $, the acceptance length $ \tau $ varies across positions due to differences in prediction difficulty for $ \mathcal{R} $. This variability, depicted in Figure~\ref{fig:acceptance_topk_analysis}(a), suggests that a fixed-depth $ \mathcal{T}_{\text{draft}} $ either overextends in regions of low $ \tau $, reducing the acceptance rate $ \alpha=\tau/\hat{d} $, or underextends in regions of high $ \tau $, constraining the expected ratio $ \mathbb{E}[\frac{\tau}{T_{\text{draft}}}] $.
	
\begin{figure}[t]
		\centering
		\includegraphics[width=1\linewidth]{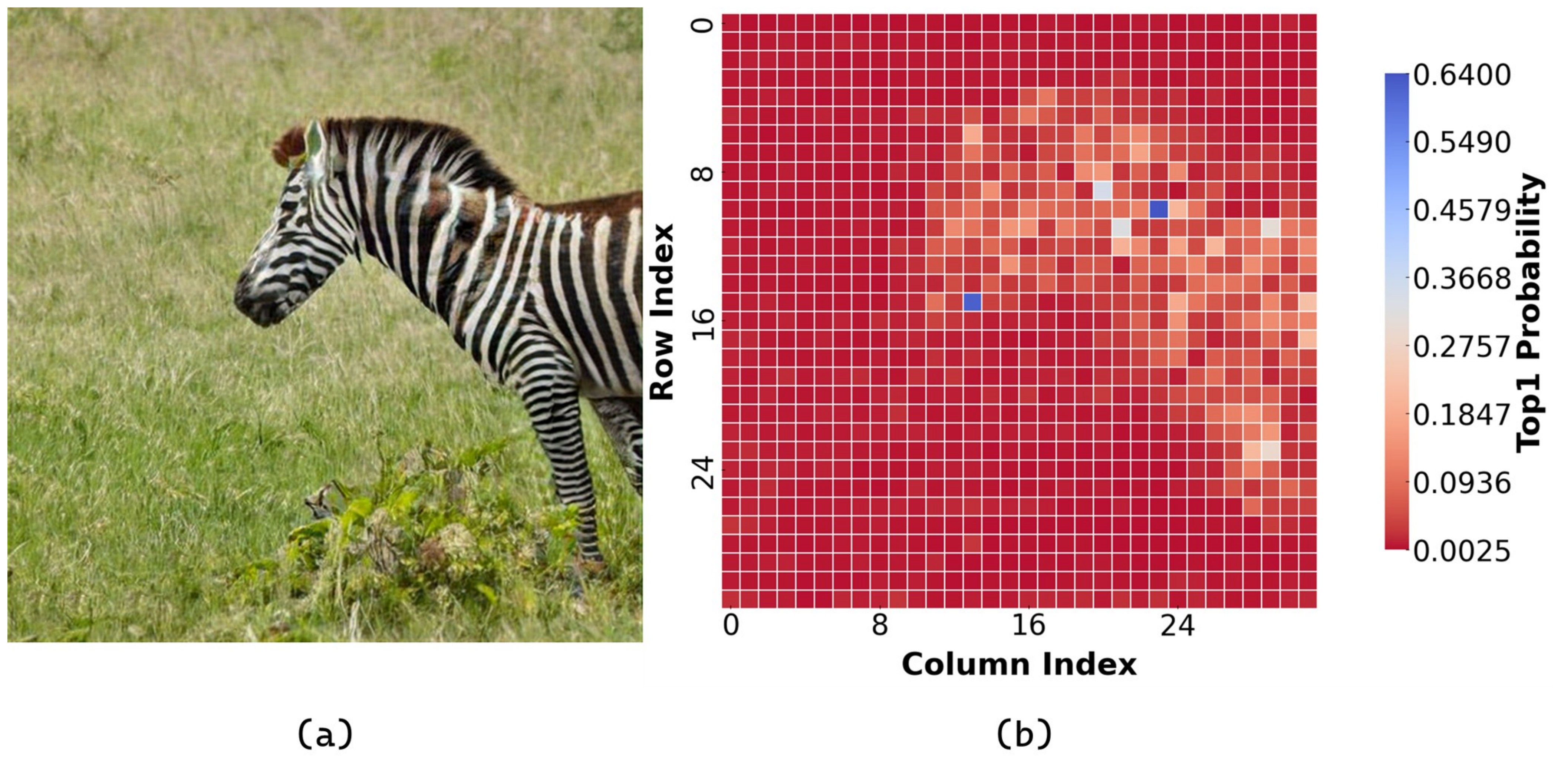}
		\caption{\textbf{(a) The original image generated from 32×32 image tokens.} \textbf{(b) The top-1 probability values of the generation distribution for each of the 32×32 image tokens.} The model is Anole, with the prompt ``a zebra``. }
		\label{fig:intro2.2}
	\end{figure}

\begin{figure}[t]
		\centering
		\includegraphics[width=1\linewidth]{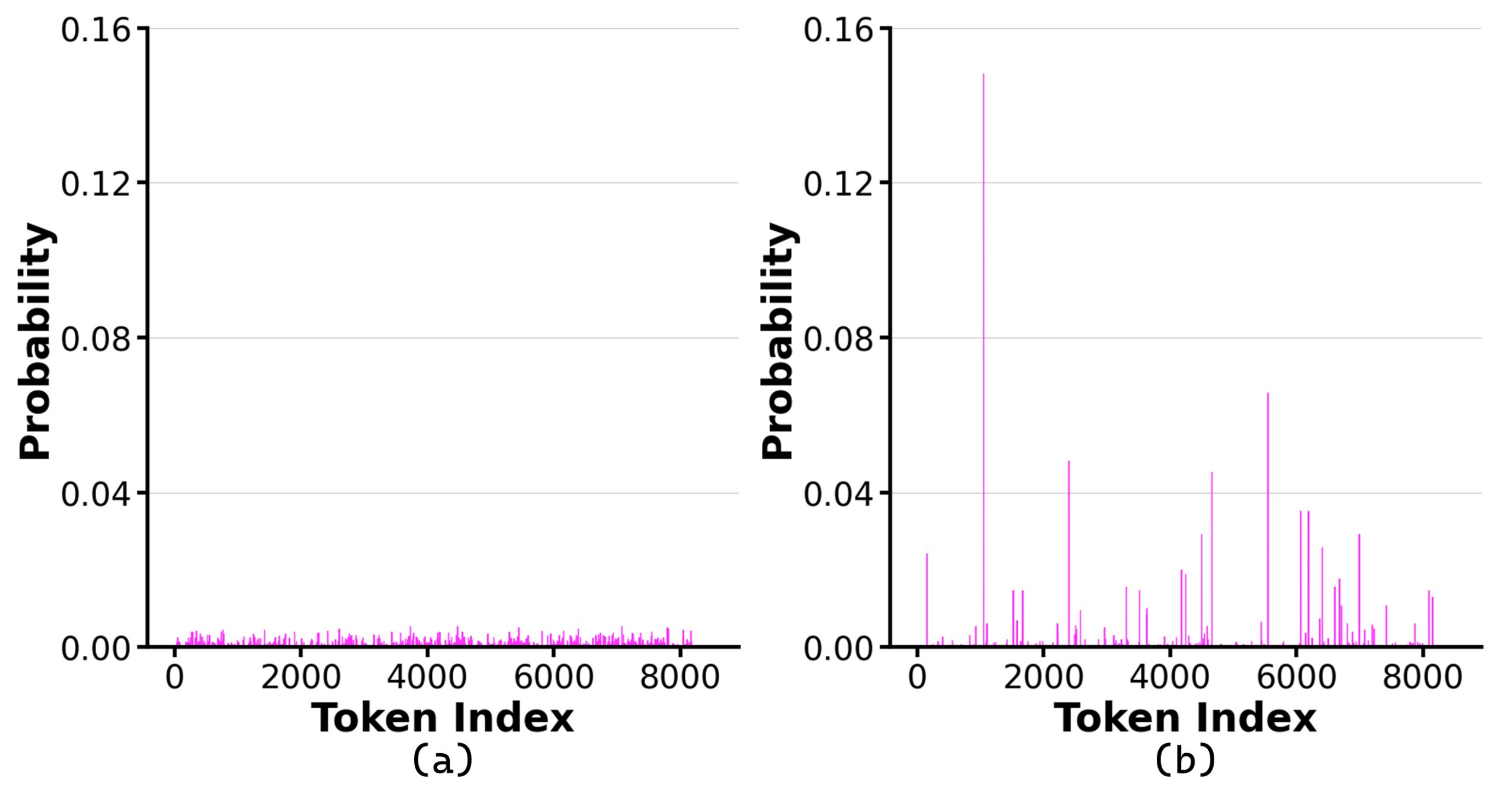}
		\caption{\textbf{(a) The probability distribution for simpler texture regions.} It corresponds to the position (8,1) in Figure~\ref{fig:intro2.2}(a). \textbf{(d) The probability distribution for complex texture regions.} It corresponds to the position (9,20) in Figure~\ref{fig:intro2.2}(b). It can be observed that the probability distribution for simpler texture regions is more uniform, with the maximum value not exceeding 0.01, whereas the probability distribution for complex texture regions is significantly more peaked, with the maximum value surpassing 0.14.}
		\label{fig:intro2.3}
	\end{figure}

	Furthermore, as shown in Figure~\ref{fig:intro2.2}(b), the root cause of the disparate acceptance rates across image regions lies in the underlying differences in generation distributions~\cite{li2015noref,guo2002synergizing}. In \textbf{simple textures regions} (e.g., low-frequency backgrounds), the top-1 probability under the draft model is often extremely low. Conversely, in \textbf{complex texture regions} (e.g., high-frequency details such as fur), the top-1 probability is typically significantly higher. As shown in Figure~\ref{fig:intro2.3}(a) and Figure~\ref{fig:intro2.3}(b), this results in simpler texture regions exhibiting a more uniform generation distribution and thus higher diversity. Conversely, complex texture regions display a more concentrated distribution, leading to lower diversity and higher certainty. 
	
	Based on the above observations, we propose a potential solution: regions with simple textures in the generated image exhibit higher $\tau$ values, as $\mathcal{R}$ can predict tokens more accurately, and when visual error tolerance is high, the distribution discrepancy between the draft model and target model is smaller. In contrast, complex texture regions show lower $\tau$ values due to reduced visual error tolerance, resulting in significant distribution divergence between $q(\cdot | s_{1:t-1}, \rho)$ and $p(\cdot | s_{1:t-1}, \rho)$. This behavior is closely related to the spatial coherence~\cite{A2DM,IESR,AIG} of images—adjacent tokens demonstrate strong correlations in acceptance lengths, reflecting local consistency in generation difficulty~\cite{robust,hawk}. Leveraging this property, we can dynamically adjust the structure of the draft tree by analyzing the acceptance rates of neighboring regions.
	
	Additionally, Figure~\ref{fig:acceptance_topk_analysis}(b) reveals variability in the top-$ k $ positions of accepted tokens within $ q(\cdot | s_{1:t-1}, \rho) $. In complex regions, the position of draft tokens’ probabilities may rank lower in $ \mathcal{R} $’s distribution compared to $ \mathcal{L} $, occasionally falling outside the top-$ k $ range ($ k_d > \hat{k} $), leading to rejection. This discrepancy underscores the need for adaptive $ \hat{k} $ alongside $ \hat{d} $.
	
	Motivated by these findings, we propose ADT-Tree, an algorithm that dynamically adjusts the depth $ \hat{d} $ and width $ \hat{k} $ of $ \mathcal{T}_{\text{draft}} $ during the expansion phase of speculative decoding. By tailoring the tree structure to the local prediction difficulty, ADT-Tree aims to maximize $ \mathbb{E}[\frac{\tau}{T_{\text{draft}}}] $ while minimizing unnecessary computation.
	
	\section{ADT-Tree: Adjacency-Adaptive Dynamical Draft Trees}\label{sec:4}
	To address the challenge of uneven acceptance rates across draft trees at various positions, stemming from inconsistent acceptance lengths, we introduce Adjacency-Adaptive Dynamical Draft Trees, dubbed ADT-Tree. This approach dynamically builds a draft tree by adapting to the acceptance rate state of adjacent tokens in visual auto-regressive models. Let $\hat{d}$ be the depth of the draft tree and $\hat{k}$ be the width (the top-k value) of the draft tree.
	ADT-Tree constructs the draft tree through two phases: initialization and adaptation. First, the depth $\tilde{d}$ and width $\tilde{k}$ of the current draft tree are initialized according to the established strategy. Second, it revises these two values according to the acceptance rate of the previous draft trees, which reflects the current level of prediction difficulty. Details are described below.
	
	\subsection{Adjacent Initialization}
	\label{sec:4.1}
	
	We introduce Adjacent Initialization, which addresses the initialization of the depth $\tilde{d}$ and top-k $\tilde{k}$ for a new draft tree associated with the image token $s^{(i,j)}$. Let $\mathbf{(i,j)}$ denote the position of the token to be predicted within a two-dimensional grid representing the encoded image. Three strategies that leverage adjacency to initialize $\tilde{d}$ and $\tilde{k}$ are provided as follows:
	\begin{itemize}
		\item \textbf{Horizontal Repeat (Repeat Left Adjacent Draft Tree)}: Set $\tilde{d} = d^{i,j-1}$ and $\tilde{k} = k^{i,j-1}$, using the draft tree attributes of $s^{(i,j-1)}$.
		\item \textbf{Vertical Repeat (Repeat Above Adjacent Draft Tree)}: Set $\tilde{d} = d^{i-1,j}$ and $\tilde{k} = k^{i-1,j}$, based on $s^{(i-1,j)}$.
		\item \textbf{Random Initialization}: Sample $\tilde{d} \sim \mathcal{U}(d_{\text{min}}, d_{\text{max}})$ and $\tilde{k} \sim \mathcal{U}(k_{\text{min}}, k_{\text{max}})$.
	\end{itemize}
	
	These three strategies build on experimental findings from SJD~\cite{SJD}, which reveal that tokens positioned next to each other horizontally or vertically exhibit similar probabilities during image generation. Building on this insight, the Horizontal Repeat and Vertical Repeat strategies were developed. Meanwhile, the Random strategy was crafted for images where the correlation between adjacent tokens is comparatively weak. These simple yet effective approaches adapt $\tilde{d}$ and $\tilde{k}$ to the spatial context of $s^{(i,j)}$.
	
	\begin{figure*}[t]
		\centering
		\includegraphics[width=1\linewidth]{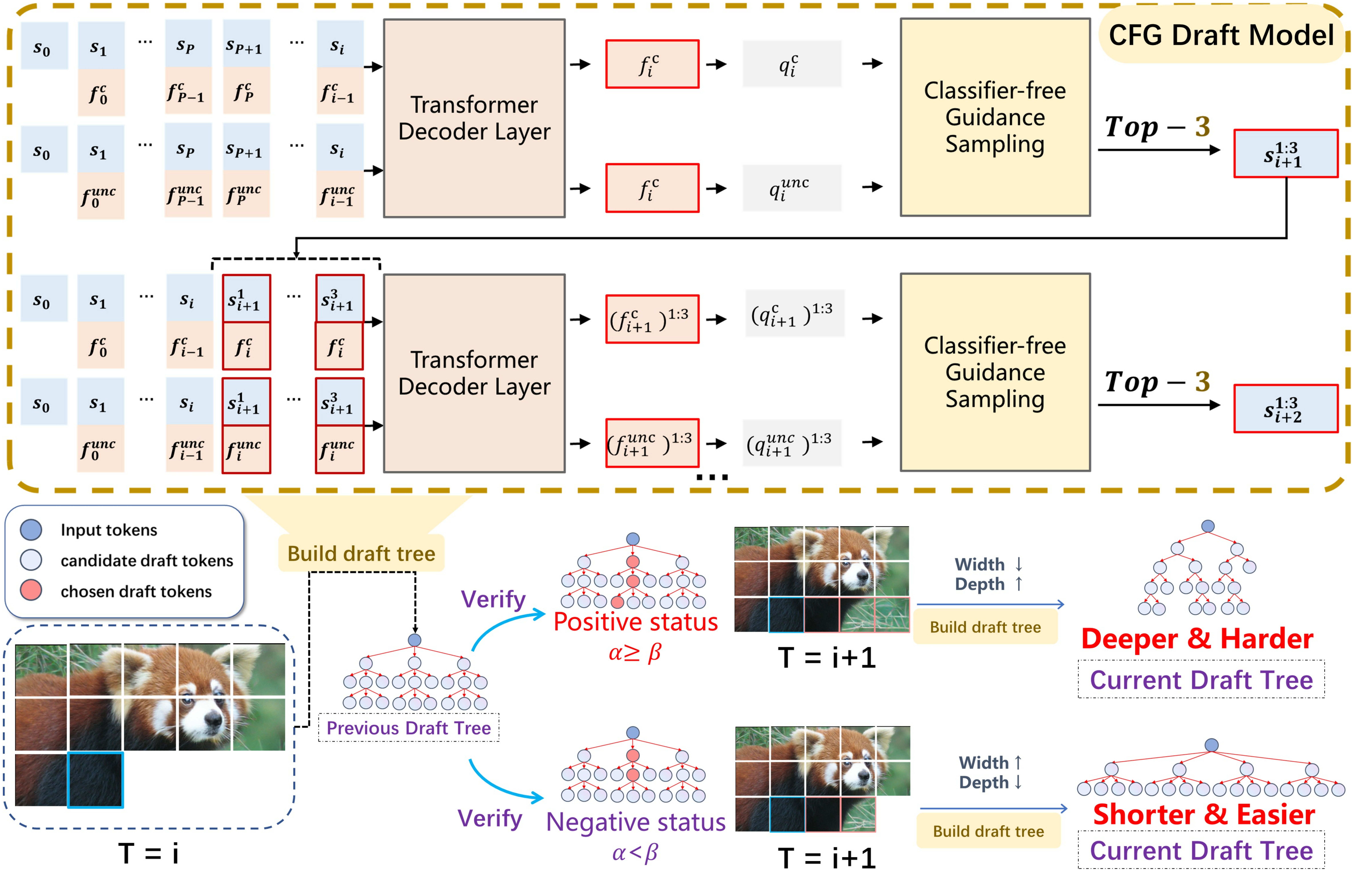}
		\caption{\textbf{The adapting phase main process.} According to the positive threshold $\beta$, the previous draft tree is classified as either a positive or a negative state. The draft tree is constructed by the draft model, whose framework is adapted from the EAGLE architecture, incorporating a forward mechanism specifically designed for classifier-free guidance~\cite{CFG}. As illustrated by the ``CFG Draft Model'' (green dashed box) in the upper part of the figure, image tokens are processed separately under conditional and unconditional inference paths. Here, $f^c_i$ and $f^{unc}_i$ denote the intermediate features at position $i$ obtained after the EAGLE FC layers~\cite{EAGLE-1} for the conditional and unconditional branches, respectively. Similarly, $q^c_i$ and $q^{unc}_i$ represent the corresponding probability distributions at position $i$ under the conditional and unconditional settings. The example in the figure is designed to construct a three-layer draft tree, thus requiring the draft model to perform three forward passes. The width of the draft tree is decided by the setting of top-k. After constructing each layer, the top-k nodes with the highest EAGLE-2 scores at that layer are selected as the parent nodes for the next layer. If it is from a complex texture to a simple texture, ADT-Tree's draft trees will turn deeper and narrower, which will remove unnecessary expansions. On the contrary, it will turn into a shorter and wider tree so that the current draft tree can more easily find the correct target tokens.}
		\label{fig:Figure5}
	\end{figure*}
	
	\subsection{Bisectional Dynamic Adaptation} \label{sec:4.2} After obtaining the initial draft tree, we design Bisectional Dynamic Adaptation to adjust the depth and width of the draft tree. As shown in Figure~\ref{fig:Figure5}, we categorize the input draft tree into two statuses. Let $\beta$ be the positive threshold, which determines the positive or negative status. 
	
	In the positive status, the acceptance rate $\alpha$ exceeds the threshold $\beta$, meaning that all tokens in each layer of the draft tree are fully utilized. In this scenario, the capability of the draft model is maximized from the perspective of depth, allowing it to construct deeper structures while relying less on top-k sampling.
	
	In the negative status, the acceptance rate $\alpha$ is less than the threshold $\beta$, indicating that some tokens in certain layers are not utilized, thus limiting the effectiveness of the draft model. Consequently, the draft tree should be shallower, and a larger top-k should be used to increase the likelihood of predicting the correct token.
	
	Therefore, the adjusted values of the depth $\hat{d}$ and the width $\hat{k}$ are computed as:
	\begin{align}
		\hat{d} = \begin{cases} 
			\tilde{d}+l_d, \: \alpha \geq \beta, \\
			\tilde{d}-l_d, \:  \alpha < \beta.
		\end{cases}
		\hat{k} = \begin{cases} 
			\tilde{k}-l_k, \:  \alpha \geq \beta, \\
			\tilde{k}+l_k, \: \alpha < \beta,
		\end{cases}
	\end{align}
	where $l_d$ and $l_k$ are the adjustment steps for the depth and width, respectively.
	
	
	Using a shallower draft tree with a larger top-k expands the larger search scope for per layer. A shallower draft tree effectively drafts tokens in areas with simple textures. Conversely, when using a deeper draft tree, reducing the top-k can minimize the space and time costs associated with building the draft tree. As. Accurately predicting the positions of child nodes containing the accepted tokens enhances the efficiency of building draft trees, thereby increasing the generation speedup.
	
	A narrower tree can also address the issue of the EAGLE draft tree becoming excessively shallow during image generation. In image generation, the distribution of image tokens is overly flat. Therefore, compared to the probability distribution of text tokens, the average generation probability of image tokens is significantly lower.
	
	While the EAGLE draft trees use the Bayesian probability of each node in the draft tree as its node score $\text{score}_{n}$, computed as follows:
	\begin{equation}
		\text{score}_{n} = \prod_{j=i}^{i+J} p_j,
	\end{equation}
	where $J$ represents the length of a draft token sequence in the candidate draft tree. The top-$N$ nodes with the highest scores are then selected to form the final draft tree.
	
	This mechanism causes a critical problem: due to the inherently low probabilities of image tokens, nodes with high scores tend to concentrate in the shallow layers of the tree. Consequently, deeper position nodes that have relatively higher individual probabilities are discarded. This phenomenon has also been observed and discussed in LANTERN++~\cite{LANTERN++}.
	
	As shown in Figure~\ref{fig:Figure5}, each node in the draft tree is not generated via a single forward pass of the draft model. Instead, the tokens at each layer of the draft tree are produced in parallel. For example, $s^{1:3}_{i+1}$ represents the first-layer nodes of a draft tree with width three, which are directly obtained from $s_i$ through a single top-3 sampling operation. Subsequently, $s^{1:3}_{i+1}$ is concatenated with the parent node's intermediate feature $f_i$ to infer its child nodes $(s^{1:3}_{i+2})^1$, $(s^{1:3}_{i+2})^2$, and $(s^{1:3}_{i+2})^3$, collectively denoted as $(s^{1:3}_{i+2})^{1:3}$, where $f_i$ is the intermediate feature of $s_i$ within the EAGLE framework~\cite{EAGLE-1}. Then, from these child nodes at this layer, the top 3 with the highest scores are selected for the next inference step.
	Therefore, according to EAGLE-2~\cite{EAGLE-2}, the time costs $C_{T}$ and the space peak costs $C_{S}$ of building dynamic draft trees are calculated as:
	\begin{align}
		C_{T} = T_{S} \cdot \hat{d} + T_{N} \cdot N,  \label{eq:Eagle time tree} \\
		C_{S} = \hat{k}^2 \cdot (\hat{d}-1) + \hat{k}, \label{eq:Eagle space tree}
	\end{align}
	where $N$ is the total number of tokens, $T_{S}$ denotes the time of inference of draft model, and $T_{N}$ denotes the time of building tree mask.
	
	Considering the worst-case time complexity when the current token remains in the negative state and cannot return to the positive state. Meanwhile, if the negative state suddenly transitions back to the positive state, the top-k value may become excessively large. Therefore, to prevent excessive $C_{T}$ and $C_{S}$,  we impose the constraints $d_{min}< \hat{d} <d_{max}$ and $k_{min}< \hat{k} <k_{max}$ to limit the depth and width of the draft tree. Additionally, we only restrict the top-k when $\tilde{d}\pm l_{d} >1$, as in this case, the top-k does not increase proportionally to the square of the difference.
	
	\section{Experiments}
	We conduct evaluations across the tasks of text-conditional image generation.
	\subsection{Experimental Settings}
	\label{sec:5.1}
	
\begin{table*}[tbp]
  \centering
  \caption{\textbf{The evaluation on the validation set of MSCOCO2017.} Speedup ratio is denoted by \( SR \), the mean acceptance length by \( \tau \), the mean draft tree depth by \( \bar{d} \), and the temperature by \( T \).}
  \resizebox{\linewidth}{!}{%
    \begin{tabular}{l|lrr|rr|lrr|rr}
      \toprule
            & \multicolumn{5}{c|}{T=0}              & \multicolumn{5}{c}{T=1} \\
      \cmidrule{2-11}
      Method & \multicolumn{3}{c|}{Acceleration} & \multicolumn{2}{c|}{Image Quality} & \multicolumn{3}{c|}{Acceleration} & \multicolumn{2}{c}{Image Quality} \\
      \cmidrule{2-11}
            & \multicolumn{1}{c}{SR (↑)} & \multicolumn{1}{c}{$\tau$ (↑)} & \multicolumn{1}{c|}{$\bar{d}$} & \multicolumn{1}{c}{HPSv2 (↑)} & \multicolumn{1}{c|}{CLIP Score (↑)} & \multicolumn{1}{c}{SR (↑)} & \multicolumn{1}{c}{$\tau$ (↑)} & \multicolumn{1}{c|}{$\bar{d}$} & \multicolumn{1}{c}{HPSv2 (↑)} & \multicolumn{1}{c}{CLIP Score (↑)} \\
      \midrule
      Anole~\cite{Anole} & 1.00× & 1.00  & 1.00  & 0.2309 & 0.3086 & 1.00× & 1.00  & 1.00  & 0.2360 & 0.3042 \\
      EAGLE-2~\cite{EAGLE-2} & 1.62× & 2.91  & 5.00  & 0.2338 & 0.3078 & 0.76× & 1.11  & 5.00  & 0.2361 & 0.3047 \\
      LANTERN~\cite{LANTERN} & 3.03× & 4.25  & 5.00  & 0.2188 & 0.2955 & 1.38× & \textbf{2.00} & 5.00  & 0.2303 & 0.3005 \\
      \textbf{ADT-Tree} & 2.21× & 3.40  & 3.86  & 0.2331 & 0.3081 & 1.06× & 1.10  & 2.09  & 0.2367 & 0.3047 \\
      \textbf{ADT-Tree+LANTERN} & \textbf{3.13×} & \textbf{4.86} & 5.15  & 0.2191 & 0.2965 & \textbf{1.53×} & 1.87  & 2.10  & 0.2331 & 0.3016 \\
      \bottomrule
    \end{tabular}%
  }
  \label{tab:Table1}%
\end{table*}

\begin{table*}[tbp]
  \centering
  \caption{\textbf{The evaluation on the validation set of parti-prompts.} Speedup ratio is denoted by \( SR \), the mean acceptance length by \( \tau \), the mean draft tree depth by \( \bar{d} \), and the temperature by \( T \).}
  \resizebox{\linewidth}{!}{%
    \begin{tabular}{l|lrr|rr|lrr|rr}
      \toprule
            & \multicolumn{5}{c|}{T=0}              & \multicolumn{5}{c}{T=1} \\
      \cmidrule{2-11}
      Method & \multicolumn{3}{c|}{Acceleration} & \multicolumn{2}{c|}{Image Quality} & \multicolumn{3}{c|}{Acceleration} & \multicolumn{2}{c}{Image Quality} \\
      \cmidrule{2-11}
            & \multicolumn{1}{c}{SR (↑)} & \multicolumn{1}{c}{$\tau$ (↑)} & \multicolumn{1}{c|}{$\bar{d}$} & \multicolumn{1}{c}{HPSv2 (↑)} & \multicolumn{1}{c|}{CLIP Score (↑)} & \multicolumn{1}{c}{SR (↑)} & \multicolumn{1}{c}{$\tau$ (↑)} & \multicolumn{1}{c|}{$\bar{d}$} & \multicolumn{1}{c}{HPSv2 (↑)} & \multicolumn{1}{c}{CLIP Score (↑)} \\
      \midrule
      Anole~\cite{Anole}& 1.00× & 1.00  & 1.00  & 0.2100 & 0.2731 & 1.00× & 1.00  & 1.00  & 0.2360 & 0.3089 \\
      EAGLE-2~\cite{EAGLE-2} & 1.98× & 3.57  & 5.00  & 0.2113 & 0.2744 & 0.80× & 1.26  & 5.00  & 0.2360 & 0.3084 \\
      LANTERN~\cite{LANTERN} & 2.82× & \textbf{4.46} & 5.00  & 0.2036 & 0.2663 & 1.90× & \textbf{2.08} & 5.00  & 0.2279 & 0.3029 \\
      \textbf{ADT-Tree} & 2.24× & 2.79  & 3.43  & 0.2109 & 0.2741 & 1.57× & 1.17  & 2.16  & 0.2370 & 0.3104 \\
      \textbf{ADT-Tree+LANTERN}& \textbf{3.05×} & 3.97  & 4.31  & 0.2041 & 0.2664 & \textbf{2.20×}& 1.78  & 2.70  & 0.2304 & 0.3046 \\
      \bottomrule
    \end{tabular}%
  }
  \label{tab:Table2}
\end{table*}

	\textbf{Datasets:}
	For the text-conditional image generation, we conduct experiments on the acceleration effect on parti-prompts~\cite{parti-prompts} and MS-COCO2017~\cite{MSCOCO}. We utilize random 100 captions sampling from the MS-COCO2017 validation captions to evaluate the actual speedup. The same experimental setting is also conducted for Parti-Prompts. 
	
	\noindent 
	\textbf{Evaluation Metrics:} ADT-Tree is a lightweight acceleration method that neither fine-tunes the target visual AR Models’ weights during training nor relaxes the acceptance conditions during decoding. Thus, the generation results remain unchanged in image quality as a result of the framework of EAGLE-2~\cite{EAGLE-2}. To measure the acceleration performance~\cite{foolad2017model}, we adopt the following metrics: 
	\begin{itemize}
		\item \textbf{Speedup Ratio (SR)}: The actual test speedup ratio relative to vanilla visual auto-regressive decoding.
		\item \textbf{Acceptance Length ($\tau$)}: The average number of tokens generated per drafting-verification cycle, indicating the number of tokens accepted by the target visual AR Model decoding from the draft model.
		\item \textbf{Mean Draft Trees Depth ($\bar{d}$)}: The average depth of draft trees per drafting-verification cycle, indicating the depth of draft trees by the draft model.
	\end{itemize}\bigskip
	
	\noindent \textbf{Implementation Details:} We set all generation latent size to 576 and classifier-free guidance score to 3.0~\cite{Anole}. To ensure consistency and comparability with EAGLE-2, we set temperature $T \in \{0.0, 1.0\}$. To validate our method ADT-Tree, for Anole's draft model, we set $\beta=1$ $l_d=1$, and $l_k=3$, where the depth and top-k of draft trees are limited in $(0, 10)$ and $(3, 14)$. We evaluate our approach on two different models, which are LlamaGen~\cite{Llamagen} and Anole~\cite{Anole}. For LlamaGen's draft model, we set $\beta=1$, $l_d=1$ and $l_k=10$, where the depth and top-k of draft trees are limited in $(0, 10)$ and $(3, 40)$. We evaluate each method in both the greedy decoding setting with T = 0 and the speculative decoding with T = 1.
	
	\noindent \textbf{Training Implementation:} Our training implementation is based on the open source repository of EAGLE-2. To train the text-condition draft model, we randomly sample 200k text-image pairs in LAION-COCO~\cite{LAION-COCO} dataset for Anole's draft model, which is used to train LlamaGen-XL(stage I)~\cite{LAION-COCO} target model. For Anole's draft model, we directly utilize the draft models that are already available in the LANTERN~\cite{LANTERN} project. Since LlamaGen~\cite{Llamagen} uses classifier-free guidance~\cite{CFG} to generate images, we randomly drop $10\%$ conditional embedding during training, consistent with target model training.
	
	
	\subsection{Results of Accelerated Image Generation}
	\label{sec:5.2}
	
	Table~\ref{tab:Table1} demonstrates that ADT-Tree achieves substantial acceleration compared to other methods in Anole. At a temperature of 0, ADT-Tree achieved a speedup ratio of 2.21 on MSCOCO2017, while ADT-Tree+LANTERN achieves a speedup ratio of 3.13. At a temperature of 1, ADT-Tree+LANTERN also obtains speedup ratios of 1.53.
	Among them, ADT-Tree+LANTERN refers to ADT-Tree employing LANTERN's relaxed sampling for image generation. Furthermore, it can be observed that although the acceptance length $\tau$ of our method is not always the largest, its average $\bar{d}$ is smaller than that of other methods. This is precisely the result of ADT-Tree dynamically constructing the draft tree based on image characteristics, which saves time in building the draft tree.
	
	Table~\ref{tab:Table2} further highlights the performance of ADT-Tree and ADT-Tree+LANTERN, focusing on the efficiency of the draft tree construction. Although the acceptance length $\tau$ of our method is not always the largest, its average $\bar{d}$ is smaller than that of other methods. This efficiency stems from ADT-Tree’s ability to dynamically construct the draft tree based on image characteristics, which reduces the time required for building the draft tree, thereby contributing to the observed acceleration in image generation.
	
	\begin{figure*}[tbp]
		\centering
		\includegraphics[width=1\linewidth]{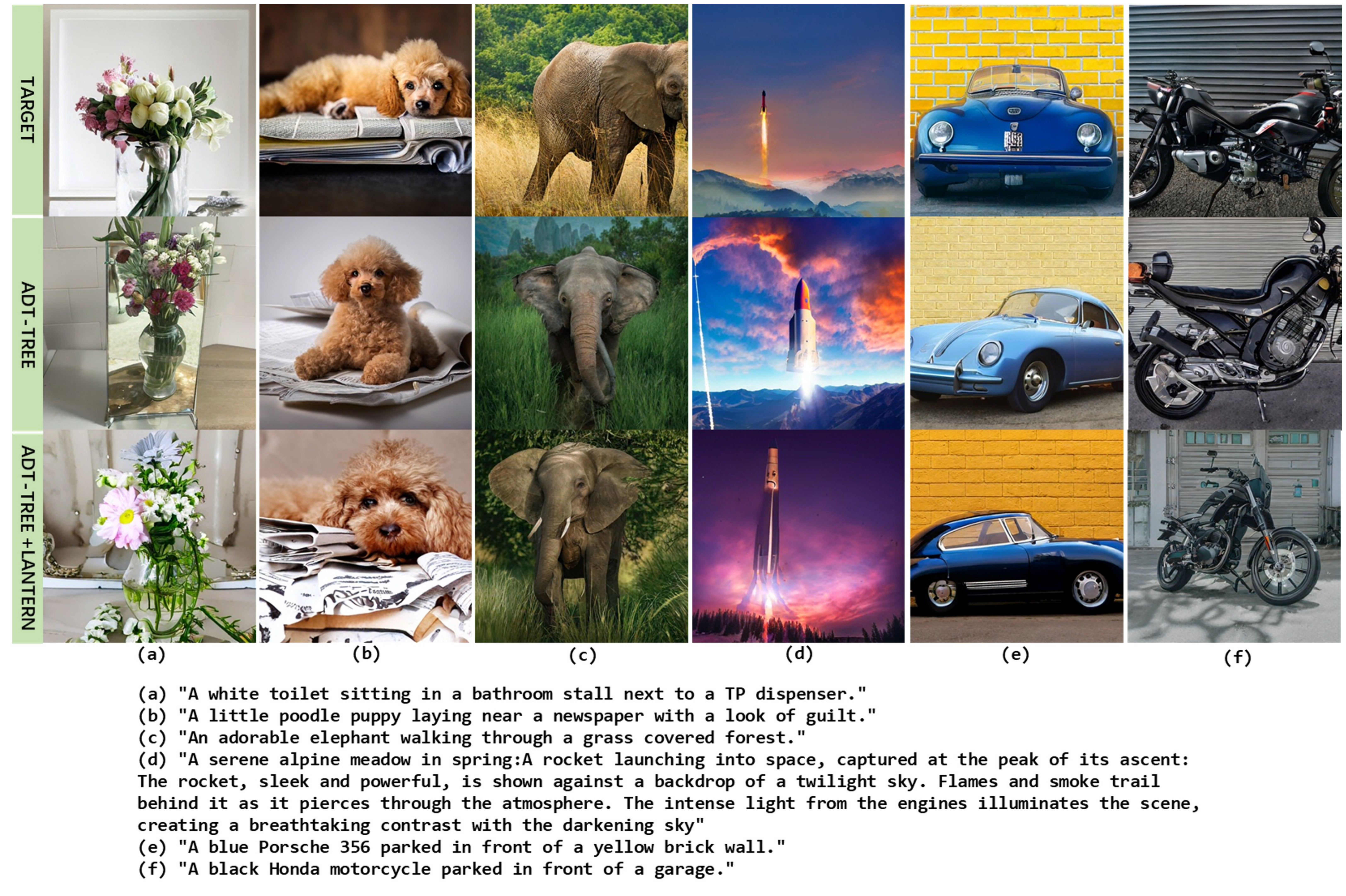} 
		\caption{\textbf{Qualitative samples generated by Anole using ADT-Tree and standard autoregressive decoding are showcased.} From top to bottom, the images correspond to outputs from standard autoregressive decoding, ADT-Tree (with parameters $l_d=1$, $l_k=3$, $\hat{d}=(0, 10)$, $\hat{k}=(3, 14)$), and ADT-Tree+LANTERN (with $\delta=0.4$, $k=1000$).
		}
		\label{fig:Figure6}
	\end{figure*}
	
	We evaluate the generated results using various image metrics.
	CLIP Score~\cite{clipscore} and HPSv2~\cite{HPSv2} measure the alignment quality between images and text.
	It can be observed that, under the same sampling methods (i.e., EAGLE-2's lossless sampling and LANTERN's relaxed sampling), ADT-Tree does not compromise the original sampling distribution. Figure~\ref{fig:Figure6} shows some images and the corresponding prompt words.

	\begin{figure}[tbp]
		\centering
		\includegraphics[width=1\linewidth]{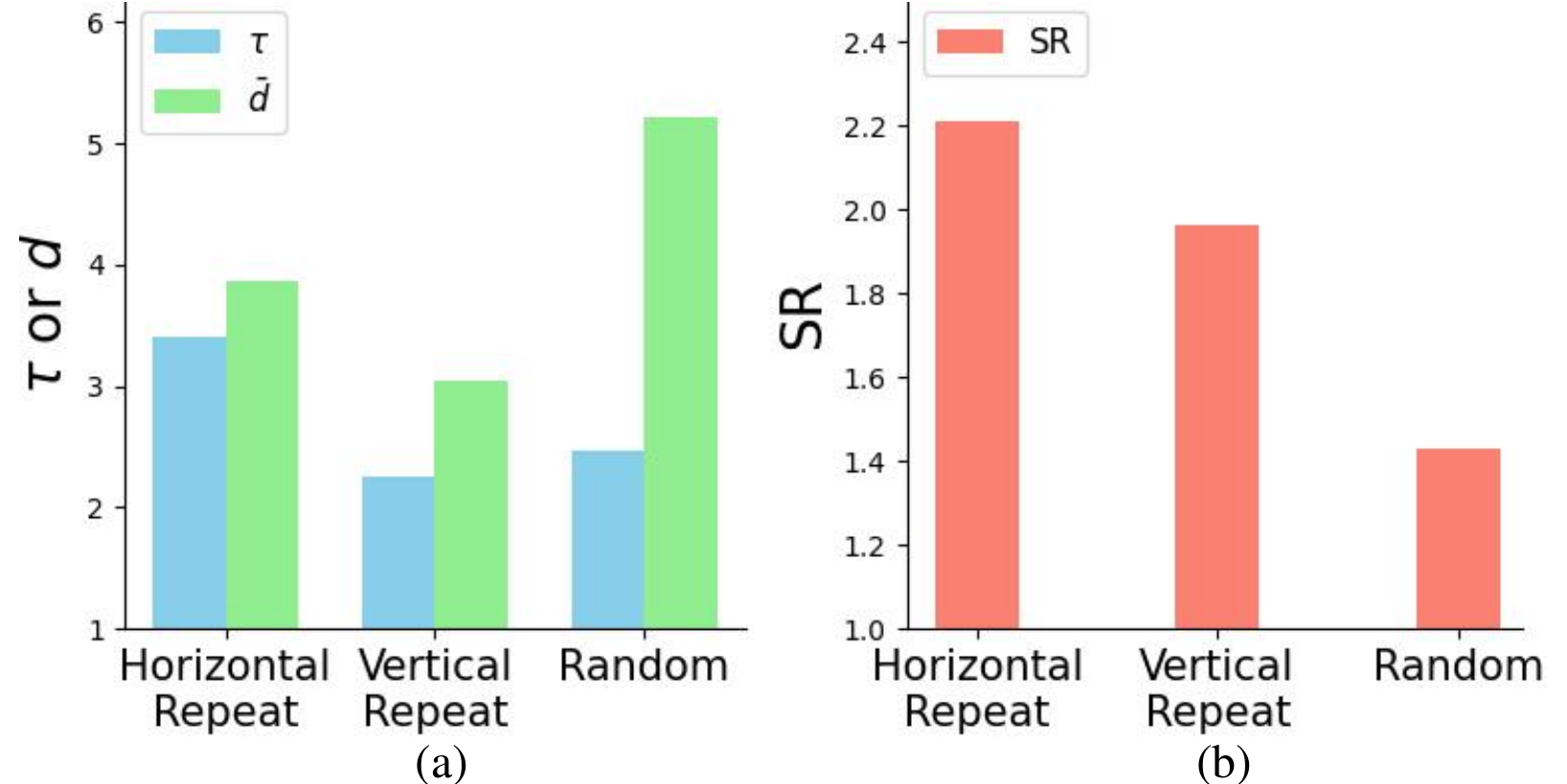}
		\caption{The influence of different initialization strategies of draft trees on the mean acceptance length $\tau$, mean draft trees depth $\bar{d}$, and speedup ratio $SR$. It can be observed that the Horizontal Repeat strategy achieves a higher speedup ratio.}
		\label{fig:Figure7}
	\end{figure}
	
	\begin{figure*}[tbp]
		\centering
		\includegraphics[width=1\linewidth]{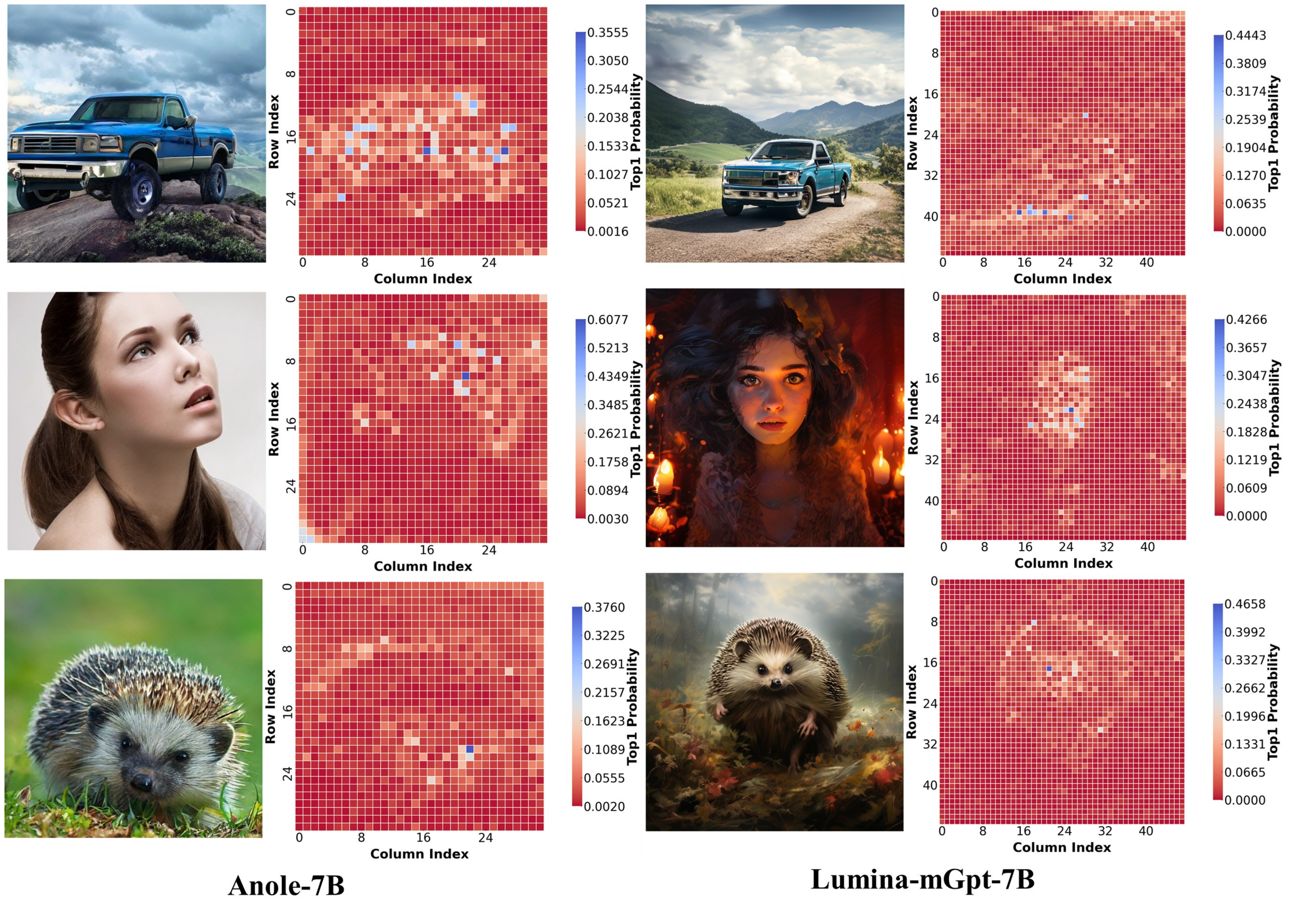} 
		\caption{\textbf{The generation results and corresponding top-1 probability distributions for Anole and Lumina-mGPT-7B-768.} The prompts from top to bottom are: ``a blue pickup truck driving in the mountains,'' ``a girl,'' and ``a hedgehog.'' It can be clearly observed that regions at object edges and those with strong semantic relevance exhibit higher top-1 probabilities, resulting in sharper generation distributions in such areas.}
		\label{fig:expcomtop1}
	\end{figure*}
	
\begin{table}[h]
\centering
\caption{\textbf{The evaluation on the validation set of MSCOCO2017.} Speedup ratio is denoted by \( SR \), the mean acceptance length by \( \tau \), the mean draft tree depth by \( \bar{d} \), and the temperature by 1.0.}
\resizebox{1\linewidth}{!}{%
\begin{tabular}{l|rrr}
\toprule
Method  &\multicolumn{3}{c}{Image Quality}\\
\cmidrule{2-4}
& \multicolumn{1}{c}{IS (↑)} & \multicolumn{1}{c}{Aesthetic (↑)} & \multicolumn{1}{c}{FID (↓)} \\
\midrule
Anole~\cite{Anole} & 30.25 & 5.93282 & 20.52 \\
EAGLE-2~\cite{EAGLE-2} & 29.87 & 5.93804 & 20.45 \\
LANTERN~\cite{LANTERN} & 27.30 & 5.81699 & 23.65 \\
\textbf{ADT-Tree} & 29.54 & 5.92874 & 22.20 \\
\textbf{ADT-Tree+LANTERN} & 28.25 & 5.86109 & 22.36 \\
\bottomrule
\end{tabular}%
}
\label{tab:Table4}%
\end{table}

\begin{table}[h]
\centering
\caption{\textbf{The evaluation on the validation set of parti-prompts.} Speedup ratio is denoted by \( SR \), the mean acceptance length by \( \tau \), the mean draft tree depth by \( \bar{d} \), and the temperature by 1.0.}
\resizebox{1\linewidth}{!}{%
\begin{tabular}{l|rr}  
\toprule
Method  & \multicolumn{2}{c}{Image Quality} \\  
\cmidrule{2-3}  
& \multicolumn{1}{c}{IS (↑)} & \multicolumn{1}{c}{Aesthetic (↑)} \\
\midrule
Anole~\cite{Anole} & 22.44 & 5.77940 \\
EAGLE-2~\cite{EAGLE-2} & 21.66 & 5.78878 \\
LANTERN~\cite{LANTERN} & 19.49 & 5.65854 \\
\textbf{ADT-Tree} & 22.34 & 5.80798 \\
\textbf{ADT-Tree+LANTERN} & 19.94 & 5.71881 \\
\bottomrule
\end{tabular}%
}
\label{tab:Table5}%
\end{table}

\begin{table}[tbp]
  \centering
  \caption{\textbf{Different calculation methods of base values' calculation results on ImageNet with Temperature=1.} L represents LlamaGen GPT-L, and XL stands for LlamaGen GPT-XL.}
  \scriptsize
  \setlength{\tabcolsep}{4pt}
  \resizebox{0.7\linewidth}{!}{%
    \begin{tabular}{clccc}
      \toprule
      Model & Method & SR & $\tau$ & $\bar{d}$ \\
      \midrule
            & TokenFlock                        & 1.32$\times$ & 2.79 & 3.60 \\
      L     & ($\tilde{d}$=1)                   & 1.27$\times$ & 2.64 & 3.23 \\
      775M  & ($\tilde{k}$=25)                  & \textbf{1.33}$\times$ & 2.84 & 3.63 \\
            & ($\tilde{d}$=1, $\tilde{k}$=25)   & 1.27$\times$ & 2.63 & 3.21 \\
            & \textbf{ADT-Tree}                   & 1.32$\times$ & \textbf{2.85} & 3.87 \\
      \midrule
            & TokenFlock                        & 1.36$\times$ & 2.52 & 3.35 \\
      XL    & ($\tilde{d}$=1)                   & 1.32$\times$ & 2.42 & 3.12 \\
      775M  & ($\tilde{k}$=25)                  & 1.36$\times$ & \textbf{2.55} & 3.37 \\
            & ($\tilde{d}$=1, $\tilde{k}$=25)   & 1.32$\times$ & 2.41 & 3.09 \\
            & \textbf{ADT-Tree}                   & \textbf{1.39}$\times$ & 2.53 & 3.51 \\
      \bottomrule
    \end{tabular}%
  }
\end{table}
	\label{tab:Phase 1}
	
	In addition, we conduct measurements of other image metrics, such as FID, IS, and Aesthetic. Table~\ref{tab:Table4} and Table~\ref{tab:Table5} demonstrate that ADT-Tree achieves acceleration without incurring any additional degradation in image generation quality. CLIP Score and HPSv2 measure the alignment quality between images and text. IS (Inception Score) is a metric for assessing the diversity and quality of generated images, Aesthetic evaluates the aesthetic quality of images, while FID (Fréchet Inception Distance) ~\cite{FID} measures the similarity between generated and real images. Since the Parti-Prompts dataset lacks real images, Table~\ref{tab:Table5} does not provide FID results. It can be observed that, under the same sampling methods (i.e., EAGLE-2's lossless sampling and LANTERN's relaxed sampling), ADT-Tree does not compromise the original sampling distribution.

	\subsection{Ablations and Analysis}
	\label{sec:5.3}
	
	\noindent \textbf{Impact of Initialization Strategy of Draft Trees: }
	We evaluate three distinct draft tree initialization strategies. As Figure~\ref{fig:Figure7} shows, the results demonstrate that the ``Horizontal Repeat" strategy achieves a significantly higher speedup ratio. As illustrated in the left sub-figure of the accompanying Figure~\ref{fig:Figure7}, the random strategy, with its random sampling approach, produces an excessively high $\bar{d}$, causing the acceleration effect to degrade.
	
	Conversely, the ``Vertical Repeat" strategy yields an initialization that is less precise than that of the ``Horizontal Repeat" strategy. This discrepancy can be attributed to the properties of the AR model: image tokens adjacent in the horizontal direction share more similar contextual information compared to those in the vertical direction, resulting in greater similarity in their acceptance lengths. Consequently, ADT-Tree ultimately adopts the ``Horizontal Repeat" strategy as its initialization approach.
	
	\noindent \textbf{Impact of Calculation Methods of Initial Values: }
	In addition to the calculation methods of $\tilde{d}$ and $\tilde{k}$, Table~\ref{tab:Phase 1} demonstrates several alternative approaches. Initially, we explore the incorporation of adjacent tokens in the position of the image patch following image token decoding, a method we designate as $TokenFlock$. It utilizes a distinct search range defined as follows: The position of a past image token $s^{(x_j, y_j)}$ is denoted by $\mathbf{(x_{j}, y_{j})}$, and the current predicted position is $\mathbf{(x_{i}, y_{i})}$. We can derive a set $\Omega$ containing positions that fall within the $\delta$-range of the current token as:
	\begin{equation}
		\Omega=\{t_{j}\mid \sqrt{(x_{i}-x_{j})^2+(y_{i}-y_{j})^2} \leq \delta,j<i\},
	\end{equation}
	where the hyper-parameter $\delta$ is used to select the top $\delta$ most adjacent tokens. We calculate the initial depth $\tilde{d}$ and initial $\tilde{k}$ by
	\begin{equation}
		\tilde{d} = \sum_{t_{j} \in \Omega} d_{j} \cdot norm\left(\frac{\delta - (y_{i} - y_{j}) + 1}{\delta}\right) 
	\end{equation}
	\begin{equation}
		\tilde{k} = \sum_{t_{j} \in \Omega} k_{j} \cdot norm\left(\frac{\delta - (y_{i} - y_{j}) + 1}{\delta}\right)
	\end{equation}
	, where $norm(\cdot)$ represents a normalization function. Subsequently, we investigate the impact of fixing specific parameters within ADT-Tree, namely $\tilde{d}$ and $\tilde{k}$, to validate the rationale behind the base value calculation method. In the table, ($\tilde{d}=1$) and ($\tilde{d}=1$, $\tilde{k}=25$) respectively represent the ADT-Tree method using specific fixed tree-building attributes. Our findings reveal that when either $\tilde{d}$ or $\tilde{k}$ is held constant, the acceleration ratio exhibits a certain degree of fluctuation. This fluctuation is particularly pronounced in the acceptance length, especially when $\tilde{d}$ is fixed.
	
	\noindent \textbf{Impact of different models on top-1 probability distributions: } We further visualize the top-1 probability values of generation distributions on Lumina-mGPT~\cite{luminamgpt} and Anole~\cite{Anole}. As shown in Figure~\ref{fig:expcomtop1}, the top-1 distributions of mainstream models consistently exhibit the phenomenon observed in Section~\ref{sec:Motivation}. We conduct experiments on object, human, and animal categories. The complex texture regions in generated images are typically semantically critical areas, such as human faces, eyes, and boundaries between objects and backgrounds. During training of AR visual generation models, these regions serve as key indicators for evaluating generation quality, and optimizers often impose stronger supervision on their reconstruction loss. Consequently, the generation distributions in these regions tend to be sharper, with higher certainty and lower diversity.
	
	Therefore, as long as an image generation model displays this characteristic distribution pattern, ADT-Tree can be effectively applied to optimize inference acceleration.
	
	\section{Conclusion and Limitations}
	In this paper, we tackle the challenge of improving inference efficiency in visual AR models. We identify a critical limitation of existing speculative decoding approaches when applied to visual AR models: the imbalance in draft tree acceptance rates caused by varying prediction difficulties across image regions. The essence of this issue lies in the varying image generation distributions across different regions, which is that simple regions exhibit more uniform distributions, whereas complex regions display sharper distributions. To address this, we propose ADT-Tree, an adjacency-adaptive method that dynamically adjusts draft tree depth and top-k based on the states of adjacent tokens and prior acceptance rates. 
	
	Experimental results on text-conditional generation tasks demonstrate that ADT-Tree dramatically outperforms baselines in inference speed while maintaining generation quality. However, one limitation of our approach lies in the fact that the module of bisectional dynamic adaptation is ineffective in cases where the generated images have essentially the same acceptance lengths. In the future, we will design visual feature-oriented adaptation modules to further enhance efficiency.
	
	\noindent\textbf{Limitations:} 
	One limitation of our approach is that when the generation distributions are highly similar across all regions, ADT-Tree may offer only marginal improvement over the baseline acceleration method.
	
	\bibliographystyle{IEEEtran}
	\bibliography{IEEEbib}
	
	
\end{document}